\documentclass[preprint,11pt]{elsarticle}
\usepackage[english]{babel}
\usepackage{amssymb,amsmath,graphicx}
\usepackage{pifont}
\usepackage{amsthm}
\usepackage{float}
\usepackage{afterpage}
\usepackage{color}
\usepackage{enumitem}
\usepackage{setspace}
\usepackage[table]{xcolor}
\usepackage{multirow}
\usepackage{microtype}
\usepackage[status=draft]{fixme}
\usepackage{subcaption}
\usepackage{adjustbox}
\usepackage{longtable}
\usepackage{rotating}
\usepackage{setspace}
\usepackage[acronym]{glossaries}
\usepackage[hidelinks]{hyperref}
\usepackage{siunitx} 
\usepackage{xcolor,colortbl}
\definecolor{Gray}{gray}{0.8}

\usepackage{url}
\usepackage{booktabs}

\usepackage{bbm}

\usepackage{xcolor}

\usepackage[boxed,linesnumbered]{algorithm2e}
\DontPrintSemicolon
\SetAlCapSkip{0.5em}
\SetKwComment{tcp}{$\triangleright$\ }{}
\SetKwRepeat{Do}{do}{while}
\SetKwFor{Repeat}{repeat}{}{}

\usepackage{tikz}
\usetikzlibrary{arrows}
\usetikzlibrary{decorations.markings}
\usepackage{pgfplots, pgfplotstable} 

\newcommand{\cmark}{\ding{51}}%
\newcolumntype{H}{>{\setbox0=\hbox\bgroup}c<{\egroup}@{}}

\newtheorem{property}{Property}

\let\oldcases\cases
\let\oldendcases\endcases
\renewenvironment{cases}{\setstretch{1}\oldcases}{\oldendcases}

\allowdisplaybreaks
\newacronym{SVM}{SVM}{Support vector machine}
\newacronym{SVM-HL}{SVM-HL}{SVM with hinge loss}
\newacronym{SVM-RL}{SVM-RL}{SVM with ramp loss}
\newacronym{SVM-HML}{SVM-HML}{SVM with hard-margin loss}
\newacronym{CB}{CB}{Combinatorial Benders'}
\newacronym{MP}{MP}{master problem}
\newacronym{SP}{SP}{subproblem}
\newacronym{MIS}{MIS}{minimal infeasible subsystem}
\newacronym{MIP}{MIP}{mixed-integer programming problem}
\newacronym{LP}{LP}{linear programming problem}
\newacronym{MIQP}{MIQP}{mixed-integer quadratic problem}

\pgfplotsset{compat=1.16}

\title{Support Vector Machines with the Hard-Margin Loss: Optimal Training via Combinatorial Benders' Cuts}

\selectfont

\begin{document}

\begin{center}

\vspace*{0.2cm}

\begin{LARGE}
Support Vector Machines with the Hard-Margin Loss:\vspace*{0.1cm}\linebreak Optimal Training via Combinatorial Benders' Cuts
\end{LARGE}
 
\vspace*{0.8cm}
 
\textbf{\'{I}talo Santana$^{1}$, Breno Serrano$^2$, Maximilian Schiffer$^{2,3}$, Thibaut Vidal$^{1,4,5}$} \\
 $^1$ Department of Computer Science, Pontifical Catholic University of Rio de Janeiro\\
 $^2$ TUM School of Management, Technical University of Munich \\
 $^3$ Munich Data Science Institute, Technical University of Munich \\
 $^4$ CIRRELT \& Department of Mathematics and Industrial Engineering, Polytechnique Montr\'eal\\
 $^5$ SCALE-AI Chair in Data-Driven Supply Chains
 
\vspace*{0.5cm}

\end{center}

\noindent
\textbf{Abstract.}
The classical hinge-loss support vector machines (SVMs) model is sensitive to outlier observations due to the unboundedness of its loss function. To circumvent this issue, recent studies have focused on non-convex loss functions, such as the hard-margin loss, which associates a constant penalty to any misclassified or within-margin sample. Applying this loss function yields much-needed robustness for critical applications but it also leads to an NP-hard model that makes training difficult, since current exact optimization algorithms show limited scalability, whereas heuristics are not able to find high-quality solutions consistently. Against this background, we propose new integer programming strategies that significantly improve our ability to train the hard-margin SVM model to global optimality. We introduce an iterative sampling and decomposition approach, in which smaller subproblems are used to separate combinatorial Benders' cuts. Those cuts, used within a branch-and-cut algorithm, permit to converge much more quickly towards a global optimum. Through extensive numerical analyses on classical benchmark data sets, our solution algorithm solves, for the first time, 117 new data sets to optimality and achieves a reduction of 50\% in the average optimality gap for the hardest datasets of the benchmark.\\

 
\noindent
$^*$ Correspondence to: \url{isantana@inf.puc-rio.br}, \url{thibaut.vidal@cirrelt.ca}

\thispagestyle{empty}

\pagebreak

\pagenumbering{arabic}

\section{Introduction}

\glspl{SVM} are among the most popular classification models due to their simplicity and solid theoretical foundations from statistical learning~\citep{Mavroforakis2006,Vapnik1998,Xu2009}. Application fields of \glspl{SVM} include, among others, image classification, bioinformatics, handwritten digits recognition, face detection, and generalized predictive control~\cite{Byun2002, Cervantes2020,Cristianini2000introduction}. Beyond this, \glspl{SVM} are regularly used as elementary building blocks of sophisticated AutoML pipelines \citep{Feurer2015}. They achieve state-of-the-art results for a variety of applications, especially for large-scale datasets~\citep{Chang2011,Collobert2001,Joachims1999}.

\begin{figure}[htbp]
 \centering
\begin{minipage}{.45\textwidth}
    \centering
    \includegraphics[width=\textwidth]{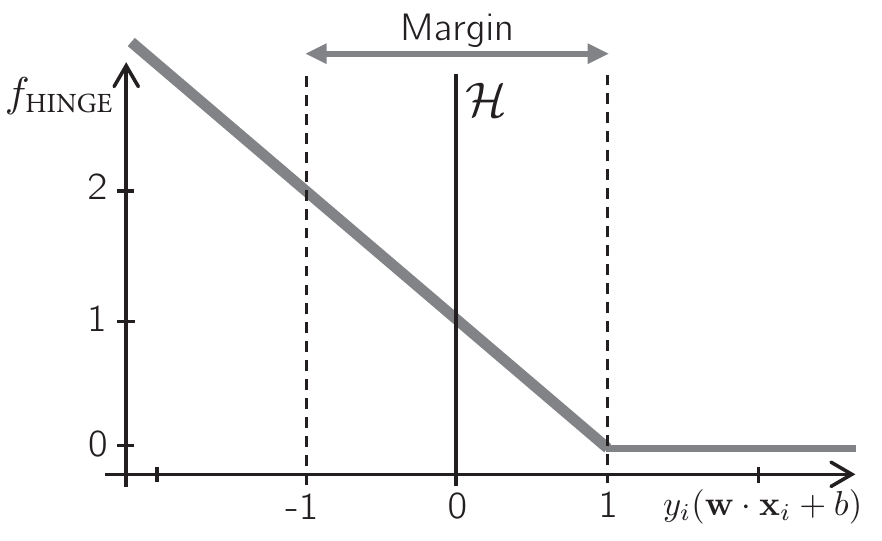}
	\caption{Hinge loss function}
	\label{fig:hinge-loss}
\end{minipage}%
\hfill
\begin{minipage}{.45\textwidth}
    \centering
    \includegraphics[width=\textwidth]{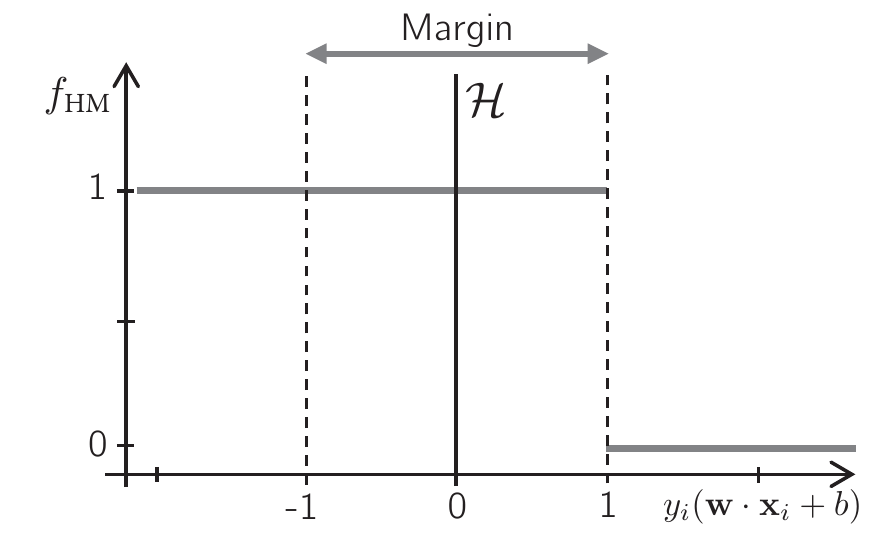}
	\caption{Hard-margin loss function}
	\label{fig:hard-margin-loss}
\end{minipage}
\end{figure}

In its simplest form, an \gls{SVM} can be defined as follows. Let $(\mathbf{X},\mathbf{y})=\{\mathbf{x}_i,y_i\}$ be a training set in which each $\mathbf{x}_i \in \mathbb{R}^m$ is an $m$-dimensional feature vector, and each $y_i \in \{-1,1\}$ is its associated class. Then, an \gls{SVM} seeks a hyperplane $\mathcal{H} = \{ \mathbf{x} \in \mathbb{R}^m : \mathbf{w} \cdot \mathbf{x} + b = 0 \}$ that optimizes the following objective:
\begin{equation}
\min\limits_{\mathbf{w},b}  \frac{1}{2} ||\mathbf{w}||^2 + C \sum_{i=1}^n f(y_i(\mathbf{w} \cdot \mathbf{x}_i + b)).
\label{general-problem}
\end{equation}
The first term of Equation~\eqref{general-problem} acts as a regularization term and indirectly maximizes the margin of the \gls{SVM}~\citep{Vapnik1998}, whereas the second term ensures fidelity to the data and penalizes misclassified samples, with coefficient $C$ balancing the two terms. Accordingly, the objective establishes a trade-off between maximizing the hyperplane's margin and minimizing the concomitant misclassification error. The loss function $f$ varies with respect to the studied problem variant. In the classical \gls{SVM-HL}, we define 
\begin{equation}
f(u) := f_\textsc{Hinge}(u) = \max \{0,1-u\},\label{eq:hinge-loss}
\end{equation}
such that a misclassified sample, i.e., a sample for which $y_i(\mathbf{w} \cdot \mathbf{x}_i + b)<1$, directly increases the objective value of Equation~\eqref{general-problem} proportionally to its error (see Figure~\ref{fig:hinge-loss}). 

However, while the classical convex \gls{SVM-HL} permits fast training and scalability, it is also known to lack robustness in the presence of misclassified samples or outliers since its loss function is unbounded. As a drawback, the trained model can be severely affected by these outliers, preventing its application in several domains, especially for high-stakes decisions where robustness is critical~\citep{Rudin2019}.

In light of this, some works have considered the use of bounded but non-convex loss functions in an attempt to gain robustness \citep[see, e.g.,][]{Brooks2011,Huang2014,Wu2007}. In particular, the \gls{SVM-HML} uses
\begin{equation}
f(u) := f_\textsc{Hard}(u) =
\begin{cases}
1 & \text{ if }  u < 1 \\
0 & \text{ otherwise,}
\end{cases}
\label{eq:hard-margin-loss}
\end{equation}
such that any misclassified sample (or sample within the margin) leads to a unit penalty (see Figure~\ref{fig:hard-margin-loss}), therefore limiting the influence of misclassified samples on $\mathcal{H}$ and increasing classification robustness.

However, this much-needed robustness comes at the expense of computational efficiency since training \gls{SVM-HML} requires solving a \gls{MIQP} and is NP-hard~\citep{Chen1996}. This hardness, but more especially the inability to efficiently solve the \gls{SVM-HML}, limits its current use. Training \gls{SVM-HML} models to global optimality is currently only achievable for very small datasets, whereas current heuristics for training do not consistently find high-quality hyperplanes.

In this work, we contribute towards addressing those challenges and pave the way toward more efficient global optimization algorithms. We propose new mixed-integer programming approaches to train \gls{SVM-HML} models to global optimality. We exploit the problem's structure to devise efficient decomposition techniques, relying on subsets of the samples to generate \gls{CB} cuts quickly. More specifically, the contribution of this paper is fourfold:
\begin{itemize}[nosep]
\item We show that \gls{CB} cuts can be successfully exploited to generate useful inequalities for \gls{SVM-HML}.
\item We introduce sampling strategies that permit to quickly generate a diversified pool of cuts. We effectively embed these cuts within a branch-and-cut algorithm, leading to an efficient training algorithm that can achieve global optimality.
\item We conduct an extensive numerical campaign to measure the performance of our training approach and the impact of important design choices. As seen in our experiments, this algorithm significantly improves the current status-quo regarding the solution of \gls{SVM-HML}, solving for the first time 117 new data sets to optimality and achieving a reduction of 50\% in the average optimality gap over previous approaches for the hardest datasets of the benchmark.
\item Generally, our study underlines the benefits of applying cutting-edge mixed-integer programming techniques to combinatorial optimization problems that arise when training non-convex machine learning models. 
\end{itemize}

The remainder of this paper is organized as follows.
Section~\ref{section:related-literature} briefly reviews the related literature.
Section~\ref{section:methodology} introduces the proposed methodology. Section~\ref{section:experiments} details our computational experiments, and Section~\ref{section:conclusions} concludes this paper.

\section{Related Literature}
\label{section:related-literature}

A vast body of literature on \glspl{SVM} exists, covering various topics such as applications, training algorithms, and loss functions. For the sake of brevity, we focus on recent contributions to training algorithms for \gls{SVM-HL} as well as works on \glspl{SVM} with non-convex loss functions, namely \gls{SVM-HML} and the \gls{SVM-RL}.

The training problem for the classical \gls{SVM-HL} can be cast and efficiently solved as a continuous convex quadratic programming problem. Existing solution approaches typically detect and fix violations of first-order optimality conditions, leading to a series of small subproblems with few variables. A classical approach, called Sequential Minimal Optimization (SMO) is used in several state-of-the-art implementations \cite{Chang2011, Collobert2001, Joachims1999,Platt1998} and consists of solving a restricted problem of only two variables at each iteration. Still, some algorithms have also exploited larger subproblems \citep[see, e.g.,][]{Fan2005,Joachims1999,Vidal2019a}. 
Training algorithms for \gls{SVM-HL} are quite diverse and base on data-selection concepts \cite{lee2007reduced, loosli2007training}, geometric methods \cite{Mavroforakis2006} and heuristics \cite{Wang2004}. For a detailed presentation of algorithms for the \gls{SVM-HL} and its variants, we refer the reader to the surveys of \citet{Carrizosa2013, Cervantes2020}, and \citet{Wang2020}, as well as to the book of \citet{Scholkopf2002}.

Despite its widespread use, the sensitivity of \gls{SVM-HL} to outliers has regularly raised obstacles when dealing with critical applications. Consequently, a part of recent research has explored the possibility of using non-convex loss functions to gain robustness. \citet{Brooks2011} focused on two non-convex loss functions in particular: the \gls{SVM-HML}~\cite{Orsenigo2003, Orsenigo2004,Perez-Cruz2003} and the \gls{SVM-RL}~\cite{Collobert2006,Shawe2004,Shen2003}. Both of these functions are bounded, such that the contribution of each sample to the objective is limited. In the \gls{SVM-RL} model, any sample within the margin receives a linear penalty proportional to its distance to the margin (a value between $0$ and $2C$), whereas any misclassified sample outside the margin receives a fixed penalty of $2C$. In the \gls{SVM-HML}, the penalty of any misclassified or within-margin sample is simply fixed to a constant $C$.

The solution algorithm proposed by \citet{Brooks2011} solves the training problem as a mixed-integer quadratic programming (MIQP) using state-of-the-art branch-and-cut solvers. For the \gls{SVM-HML} and \gls{SVM-RL}, the authors rely on indicator constraints representing logical implications between a binary variable representing the status of each sample (misclassified or not) and a linear constraint that evaluates its relative position from the separating hyperplane. However, it is well known that such constraints can be reformulated in linear form using a ``big-M'' constant, but doing so without carefully tuning the value of the M constant typically leads to an ineffective formulation with a weak linear relaxation, impeding an efficient solution by branch-and-cut \citep{Wolsey2020}.

For the \gls{SVM-RL} setting, \citet{Huang2014} explored the ramp loss function with $\ell_1$-penalty, resulting in a piecewise linear programming problem. Later, \citet{Belotti2016} compared different formulations for the logical constraints and concluded that aggressive bound-tightening techniques are necessary for a successful solution approach. The strategy derived from their studies has been since implemented as a standard routine for handling such constraints in the commercial solver CPLEX for MILP/MIQPs. Finally, \citet{Baldomero-Naranjo2020} tightened the M constants by solving sequences of continuous problems and Lagrangian relaxations. For the \gls{SVM-HML}, \citet{Poursaeidi2014} proposed hard-margin loss formulations within the context of multiple-instances classification, a setting in which class labels are defined as sets. Finally, in \cite{Huang2019}, the hard-margin loss was transformed into a re-scaled hinge loss function for imbalanced noisy classification.

Concluding, only a few studies have attempted to improve the state-of-the-art training algorithms for \gls{SVM-HML} after the seminal work of~\citet{Brooks2011}, often by concentrating on the handling of the logical constraints and the proper calibration of the $M$ constants. Despite this progress, optimal training remains limited to data sets counting a few hundred samples. To improve this status-quo, we investigate a different approach, which consists of the separation of \gls{CB} cuts and their combination with classical logical constraints to achieve a valid problem formulation with a better linear relaxation. Moreover, to improve computational efficiency, we rely on sampling techniques for a fast generation of diverse cuts.

\section{Methodology}
\label{section:methodology}

We first recall the classical mathematical programming formulations for the \gls{SVM} variants that are of interest for our study, namely \gls{SVM-HL} and \gls{SVM-HML}, and then proceed with a description of our algorithmic approach.

\subsection{Descriptive formulations}
\label{sec:descriptive-formulations}

\noindent
The \emph{SVM with hinge loss} (also known as \textit{soft-margin} \gls{SVM}) associates to misclassified and within-margin samples a penalty that is proportional to their distance in relation to the ``correctly classified margin'' of the found hyperplane (see Figure~\ref{fig:hinge-loss}). 

Let $\mathbf{w} \in \mathbb{R}^m$ be a vector of real variables that represents the coordinates of the hyperplane, let $b$ be its intercept to the origin, and let $\xi_i$ represent the misclassification penalty of sample~$i$. With these notations, the training problem for \gls{SVM-HL} can be mathematically formulated as:
\begin{align}
\min\limits_{\mathbf{w},b,\xi} \hspace*{0.3cm} & \frac{1}{2} ||\mathbf{w}||^2 + C \sum_{i=1}^n \xi_i  \label{obj:svmhl}\\
\text{s.t.} \hspace*{0.3cm} & y_i (\mathbf{w} \cdot \mathbf{x}_i + b ) \geq 1 - \xi_i, \quad i \in \{1,\dots,n\} \label{cons:svmhl}\\
& \xi_i \geq 0, \quad i \in \{1,\dots,n\}.\label{cons:svmhldomain}
\end{align}

Objective \eqref{obj:svmhl} seeks a maximum margin separator through the term $\frac{1}{2} ||\mathbf{w}||^2$ and minimizes the total misclassification penalty $C\sum_{i=1}^n \xi_i$, where $C > 0$ is a constant that balances both parts of the objective. Moreover, Constraints~\eqref{cons:svmhl} calculate the misclassification penalty of each sample.\\

\noindent
In the
\emph{SVM with hard-margin loss}, misclassified samples are associated a fixed penalty cost. Binary variables $z$ are used to indicate whether a sample~$i$ is misclassified or within the margin ($z_i=1$), or correctly classified ($z_i=0$). The \gls{SVM-HML} can be formulated as follows:
\begin{align}
\min\limits_{\mathbf{w},b,z} \hspace*{0.3cm} & \frac{1}{2} ||\mathbf{w}||^2 + C\sum_{i=1}^n z_i \label{obj:svmhml}\\
\text{s.t.} \hspace*{0.3cm} & (z_i = 0) \Rightarrow y_i (\mathbf{w} \cdot \mathbf{x}_i + b ) \geq 1 & i \in \{1,\dots,n\}  \label{cons:svmhml}\\
& z_i \in \{0,1\} & i \in \{1,\dots,n\}. \label{cons:svmhmldom}
\end{align}

In this formulation, the penalty term in the objective \eqref{obj:svmhml} is directly proportional to the number of misclassified samples. Constraints \eqref{cons:svmhml}, represented as logical constraints, ensure that $z_i=0$ only if sample $i$ is correctly classified. These constraints are strictly speaking not part of the semantic of a MILP/MIQP, but they could be directly transformed into a linear constraint by using a big-M constant and imposing $y_i (\mathbf{w} \cdot \mathbf{x}_i + b ) \geq 1 - M z_i$. The drawback of such a reformulation is that it leads to a formulation that provides notably weaker linear-relaxation bounds, rendering branch-and-bound algorithms relatively inefficient.

\subsection{Combinatorial Benders' cuts}
\label{subsec:cb-cuts}

To optimally solve the \gls{SVM-HML}, we propose a solution method based on Benders' decomposition~\cite{Benders1962}. In its canonical form, this strategy exploits the structure of a mixed-integer linear program and splits its variables into two subsets. The first subset of integer or continuous variables (sometimes called ``complicating variables'') is selected in such a way that fixing them either decomposes or reduces the complexity of the resulting problem. The remaining variables should be continuous. The method then works by decomposing the original problem into a \gls{MP}, solved over the complicating variables, and a \gls{SP}, solved as a linear program over the remaining continuous variables. The algorithm iteratively produces an incumbent solution of the \gls{MP} and uses the dual of the \gls{SP} to assess its feasibility. If the \gls{SP} determines that the incumbent solution is feasible, then this information is integrated into the \gls{MP} in the form of an additional \gls{CB} cut, which eliminates the infeasible incumbent solution.

The studies of \citet{Hooker2003} and \citet{Codato2006} paved the way towards efficient applications of Benders' decomposition to a broad class of \glspl{MIP} with logical constraints. Notably, the so-called \gls{CB} approach leads among others to a new solution paradigm for models of the following form:
\begin{align}
\min \hspace*{0.3cm} & \mathbf{c}^\intercal \mathbf{z} \label{modelCBbegin} \\
\text{s.t.} \hspace*{0.3cm}
& (z_{i} = 0) \Rightarrow \mathbf{a_i}^\intercal \, \mathbf{y} \geq d_i & i \in \{1,\dots,n\} \label{implicationCB} \\
&  \mathbf{z} \in \{0,1\}^n \\
&  \mathbf{y} \in Y, \label{modelCBend}
\end{align}
with binary complicating variables $\mathbf{z}$, continuous variables $\mathbf{y}$ in a polytope $Y$ (i.e., respecting a set of linear inequalities), and linear weights $\mathbf{c}$ (applied only on the coefficients of $\mathbf{z}$) to calculate the objective. In a \gls{CB} approach, Model~(\ref{modelCBbegin}--\ref{modelCBend}) is reformulated as follows:
\begin{align}
\min \hspace*{0.3cm} & \mathbf{c}^\intercal \mathbf{z} \\
\text{s.t.} \hspace*{0.3cm}
&\sum_{i \in S} z_i \geq 1  & S \in \mathcal{S}^\textsc{mis} \label{implicationCB2} \\
&  \mathbf{z} \in \{0,1\}^n,
\end{align}
where $\mathcal{S}^\textsc{mis}$ is the collection of all inclusion-minimal infeasible subsystems (MIS) of rows:
\begin{equation}
\mathcal{S}^\textsc{mis} = \left\{S \ \left\lvert \ 
\begin{aligned}
&\{ \mathbf{a_i}^\intercal \, \mathbf{y} \geq d_i \ \forall i \in S, \mathbf{y} \in Y \} \text{ is infeasible} \\
&\{ \mathbf{a_i}^\intercal \, \mathbf{y} \geq d_i \ \forall i \in \hat{S}, \mathbf{y} \in Y \} \text{ is feasible} \ \forall \hat {S} \subset S
\end{aligned}
\right.\right\}.\\
\end{equation}

Notably, this formulation no longer contains logical implications or big-M terms. However, it includes an exponential number of constraints. For this reason, the \gls{CB} approach consists of dynamically detecting and adding Constraints~(\ref{implicationCB2}). As in a classical Benders' decomposition, the method alternates between solving the master problem with only a subset of the constraints $\mathcal{S} \subset \mathcal{S}^\textsc{mis}$ found so far, and solving a subproblem to identify new violated constraints that cut the incumbent solution of the master in case of infeasibility. Finally, we observe that $\mathcal{S}^\textsc{mis}$ does not need to be restricted to ``inclusion-minimal'' subsets of rows to yield a valid formulation, but doing so significantly reduces the number of constraints in the set.

Despite its successful application in a variety of settings~\citep{Akpinar2017, Cote2014, Erdoan2015}, the aforementioned \gls{CB} framework is not directly applicable to problems with objective functions that contain both complicating variables~$\mathbf{z}$ and continuous variables~$\mathbf{y}$. This is unfortunately the case for the \gls{SVM-HML}, due to the occurrence of the continuous variables $\mathbf{w}$ and $b$ in the objective and the logical implications. To circumvent this issue, we introduce a weaker set of \gls{CB} cuts in conjunction with the original logical implication constraints to design an effective solution method. In this case, the cuts do not carry the burden of ensuring the model's validity, but nevertheless contribute to strengthening the formulation to obtain a better linear relaxation and improve the solution process.

\subsection{Combinatorial Benders' cuts and the SVM-HML}
\label{subsec:cb-cuts-svm}

We first start by describing a direct application of \gls{CB} to the SVM-HML and by analyzing its shortcomings. This would lead to the following formulation:
\begin{align}
\min\limits_{W,\mathbf{z}} \hspace*{0.3cm} & \frac{1}{2} W^2 + C\sum_{i=1}^n z_i \label{BC-first0} \\
\text{s.t.} \hspace*{0.3cm}
&\sum_{i \in S} z_i \geq 1  \hspace*{8cm} S \in \mathcal{S}_\textsc{svm-hml}^\textsc{mis}(W) \\
&  \mathbf{z} \in \{0,1\}^n \\
\text{with} \hspace*{0.3cm}
&\mathcal{S}_\textsc{svm-hml}^\textsc{mis}(W) = \left\{S \ \left\lvert \ 
\begin{aligned}
&\{  y_i (\mathbf{w} \cdot \mathbf{x}_i + b ) \geq 1  \ \forall i \in S, \ ||w||^2 \leq W^2 \} \text{ is infeasible} \\
&\{  y_i (\mathbf{w} \cdot \mathbf{x}_i + b ) \geq 1  \ \forall i \in \hat{S}, \  ||w||^2 \leq W^2 \} \text{ is feasible} \ \forall \hat {S} \subset S
\end{aligned}
\right.\right\}.\label{BC-first}
\end{align}

As seen in this formulation, $W$ appears in the objective and also characterizes the set of Benders' cuts. Unfortunately, the resulting formulation can no longer be practically solved as a MILP or MIQP due to this dependency. To remedy this issue, we leverage Property~\ref{monotonicity}.

\begin{property}
\emph{
Let $W_\textsc{ub}$ be a valid upper bound on $W$ on any optimal solution of Problem~(\ref{BC-first0}--\ref{BC-first}). Then, $\mathcal{S}_\textsc{svm-hml}^\textsc{mis}(W_\textsc{ub})$ is also set of valid inequalities for the SVM-HML.
}
\label{monotonicity}
\end{property}

\noindent
\textbf{Proof.} Consider $S \in \mathcal{S}_\textsc{svm-hml}^\textsc{mis}(W_\textsc{ub})$. Due to the definition of $\mathcal{S}_\textsc{svm-hml}^\textsc{mis}(W_\textsc{ub})$, $\{y_i (\mathbf{w} \cdot \mathbf{x}_i + b ) \geq 1  \ \forall i \in S$ with $||w||^2 \leq W_\textsc{ub}^2\}$ is an infeasible subsystem. Given that all optimal solutions of Problem~(\ref{BC-first0}--\ref{BC-first}) satisfy $W \leq W_\textsc{ub}$, then $\{y_i (\mathbf{w} \cdot \mathbf{x}_i + b ) \geq 1  \ \forall i \in S$ with $||w||^2 \leq W^2\}$ is also an infeasible subsystem, and therefore at least one sample $i$ in $S$ must be misclassified, implying that $\sum_{i \in S} z_i \geq 1$ is a valid inequality.
\\

With this property, the dependence upon parameter $W$ can be avoided as soon as a valid upper bound is known. The quality of the upper bound also impacts the strength of the valid inequalities obtained. Given an initial feasible solution of the SVM-HML problem with value $\Phi_\textsc{ub}$ (obtained, for example, with a heuristic for this problem), we can use $W_\textsc{ub} = 2\sqrt{\Phi_\textsc{ub}}$ since the two terms of the objective are positive.

Finally, we opted to further relax the \gls{CB} cuts by using $-W_\textsc{ub} \leq w_j \leq W_\textsc{ub}$ for $j \in \{1,\dots,m\}$ instead of $||w||^2 \leq W_\textsc{ub}^2$ in Equation~(\ref{BC-first}). Our experimental analyses have shown that this permitted a faster cut separation with only a limited impact on the strength of the formulation. Overall, we will use the resulting valid inequalities in combination with the original formulation and the tightened bounds on the $w_j$ coefficients, leading to the following model:
\begin{align}
\hspace*{-0.3cm} \min\limits_{\mathbf{w},\mathbf{z}} \hspace*{0.2cm} & \frac{1}{2} ||\mathbf{w}||^2  + C\sum_{i=1}^n z_i \label{BC-second-begin}  \\
\hspace*{-0.3cm} \text{s.t.} \hspace*{0.2cm} & (z_i = 0) \Rightarrow y_i (\mathbf{w} \cdot \mathbf{x}_i + b ) \geq 1  \hspace*{4.9cm} i \in \{1,\dots,n\} \label{BC-second-next} \\
&-W_\textsc{ub} \leq w_j \leq W_\textsc{ub} \hspace*{6.3cm} j \in \{1,\dots,m\} \\
&\sum_{i \in S} z_i \geq 1  \hspace*{8cm} S \in \mathcal{S}_\textsc{svm-hml}^\textsc{mis-ub} \label{BC-second-next2} \\
&  \mathbf{z} \in \{0,1\}^n \\
\hspace*{-0.2cm} \text{with} \hspace*{0.2cm}
&\mathcal{S}_\textsc{svm-hml}^\textsc{mis-ub} = \left\{S \ \left\lvert \ 
\begin{aligned}
&\{  y_i (\mathbf{w} \cdot \mathbf{x}_i + b ) \geq 1  \ \forall i \in S, \ -W_\textsc{ub} \leq w_j \leq W_\textsc{ub}  \ \forall j \} \text{ is infeasible} \\
&\{  y_i (\mathbf{w} \cdot \mathbf{x}_i + b ) \geq 1 \ \forall i \in \hat{S}, \  -W_\textsc{ub} \leq w_j \leq W_\textsc{ub} \ \forall j \} \text{ is feasible} \ \forall \hat {S} \subset S
\end{aligned}
\right.\right\} \hspace*{-0.05cm}.\label{BC-second-end}
\end{align}
With this problem formulation in mind, we will focus on our general solution approach and the separation of the \gls{CB} cuts in the following.

\subsection{General Solution Approach}
\label{subsec:general-solution-approach}

Our solution approach unfolds in three steps:
\begin{itemize}[nosep]
    \item[]\textbf{Step 1.} Finding an initial upper bound $W_\textsc{ub}$;
    \item[] \textbf{Step 2.} Solving a simplified formulation by branch-and-cut to obtain a cut set $\mathcal{S}_\textsc{svm-hml}^\textsc{mis-ub}$;
    \item[] \textbf{Step 3.} Solving Problem~(\ref{BC-second-begin}--\ref{BC-second-end}) with the set of cuts identified in the previous step.
\end{itemize}
We note that the generation of the cuts is done in a separate phase (Step 2). Our computational experiments have shown that it is more effective to generate a set of cuts beforehand, and then allow the MILP solver (CPLEX) to use its default settings when solving the complete model with these cuts, instead of dynamically providing additional cuts as the search progresses. We now proceed with a detailed description of each step of the algorithm.\\

\noindent
\textbf{Step 1 -- Initial bound.} 
We start by solving an SVM with hinge loss, cast as a continuous quadratic program through Equations~(\ref{obj:svmhl}--\ref{cons:svmhldomain}), and collect the resulting value of the variables $(\mathbf{w'},b',\xi')$ defining the hyperplane. With this hyperplane, we obtain an associated \gls{SVM-HML} solution by setting $z_i = \lceil \xi_i \rceil$ and calculate the resulting $W_\textsc{ub}$ objective value.\\

\noindent
\textbf{Step 2 -- Separation of the Combinatorial Benders' cuts.} Next, we consider Problem~(\ref{BC-second-begin}--\ref{BC-second-end}) excluding the logical Constraints~\eqref{BC-second-next} while dynamically generating Constraints~\eqref{BC-second-next2}. After excluding the logical constraints, $||\mathbf{w}||^2$ is free to take a value of $\mathbf{0}$, and therefore the first term of the objective vanishes. The resulting problem is a variant of the linear separator problem \citep{Kearns1994,Vapnik1998}, which we will only use to generate \gls{CB} cuts for the subsequent solution of the SVM-HML. To obtain the cuts, we apply a branch-and-cut scheme on the following master problem:
\begin{align}
 \textbf{Master:} \hspace*{0.5cm} \min\limits_{\mathbf{z}} \hspace*{0.2cm} & \sum_{i=1}^n z_i \label{master1}  \\
&  \sum_{i \in S} z_i \geq 1 \hspace*{6cm} S \in \mathcal{S} \\
&  \mathbf{z} \in \{0,1\}^n. \label{master3}
\end{align}
At each branching node, the linear relaxation of the master problem is solved and the set $\mathcal{I}^{\hspace*{0.03cm}0}$ of variables with values $z_i=0$ are identified. Based on this set of variables, we define the following feasibility subproblem to check if a violated \gls{CB} cut exists:
\begin{align}
\textbf{Subproblem:} \hspace*{0.5cm} & y_i (\mathbf{w} \cdot \mathbf{x}_i + b ) \geq 1  &\hspace{1cm} i \in \mathcal{I}^{\hspace*{0.03cm}0} \label{subproblem1}\\
&-W_\textsc{ub} \leq w_j \leq W_\textsc{ub} &\hspace{1cm} j \in \{1,\dots,m\}. \label{subproblem2}
\end{align}
If this subproblem admits a feasible solution, then no \gls{CB} cut needs to be added. 
Otherwise, the subproblem is infeasible and we can obtain an inclusion-minimal infeasible subsystem (MIS) of indices from CPLEX, giving us a new set of variables $\bar{\mathcal{I}}$ which can be added to $\mathcal{S}$. As seen in Algorithm~\ref{alg:master-subproblem}, this process is repeated at each node of the branch-and-cut tree until no additional cut can be found.

\begin{figure*}[htbp]
\centering
\begin{minipage}{0.9\textwidth}
\begin{algorithm}[H]
\linespread{1.15}\selectfont
\DontPrintSemicolon
\caption{Generation of Combinatorial Benders' cuts on a given node}
\label{alg:master-subproblem}
$\mathcal{S} \leftarrow \emptyset$\;
\Repeat{}
{
    $\mathbf{z} \leftarrow$ Solve the linear relaxation of Problem~(\ref{master1}--\ref{master3}) with $\mathcal{S}$\;
    $\mathcal{I}^{\hspace*{0.03cm}0}\leftarrow$ Indices such that $z_i = 0$\;
    \uIf{\emph{Problem~(\ref{subproblem1}--\ref{subproblem2}) with $\mathcal{I}^{\hspace*{0.03cm}0}$ admits a feasible solution}}{
        \textbf{break}
    }
    \uElse
    {
        $\bar{\mathcal{I}} \leftarrow$ MIS from Problem~(\ref{subproblem1}--\ref{subproblem2}) \;
        $\mathcal{S} \leftarrow \mathcal{S} \cup \bar{\mathcal{I}} $\;
    }
}
\end{algorithm}
\end{minipage}
\end{figure*}

To avoid spending excessive time on the solution of this auxiliary problem and to generate a diversified set of \gls{CB} cuts, we rely on sampling strategies. Until a maximum time limit of $T_\textsc{S}$, we randomly select subsets of $r = \min\{n/2,50\}$ samples, and apply the aforementioned branch-and-cut and cut separation methodology. Only afterward we repeat this process in a final run, considering all the samples and hot starting with all the \gls{CB} cuts already found. We stop the cut separation as soon as a time limit $T_\textsc{B}$ is attained.\\

\textbf{Step 3 -- Solution of the SVM-HML.} Finally, we proceed with solving the complete Problem~(\ref{BC-second-begin}--\ref{BC-second-end}), using the union of the \gls{CB} cuts that have been found in Step 2. These cuts are included as lazy constraints to avoid any overhead due to the number of cuts. Furthermore, we warm start from the solution found in Step 1, and use the default settings of CPLEX, except for the \textit{locally valid implied bounds} parameter, which is set to \emph{very aggressively} as suggested in \cite{Belotti2016}. The solution approach is run until it proves optimality or reaches a maximum time limit of $T_\textsc{Max}$. Note that this time limit encompasses all the steps of our method, such that it is possible to limit the total computational time and compare our method with other algorithms.

\section{Computational Experiments} 
\label{section:experiments}

We conduct extensive experimental analyses to evaluate the performance of the proposed approach, denoted as CB-SVM-HML, on a diverse set of benchmark instances. As an experimental baseline for this comparison, we rely on a reimplementation of the branch-and-cut approach of~\citet{Brooks2011}. The goal of our experiments is (i) to compare the performance of these algorithms in terms of computational time and optimality gaps, and (ii) to evaluate the impact of the \gls{CB} cuts and of the proposed search strategies based on sampling.

\subsection{Data and Experimental Setup}
\label{sec:exp-setup}

To evaluate the performance of our algorithm, we use the same benchmark as in~\citet{Brooks2011}, divided into three sets: UCI, Type A, and Type B.
The UCI set consists of 11 heterogeneous datasets from the UCI machine learning repository \citep{Dua2017}, which were preprocessed by~\citet{Brooks2011}.
Table~\ref{tab:uci-instances} lists the number of samples~($n$) and features~($m$) of these datasets.
Type-A and Type-B datasets have been constructed by \citet{Brooks2011} using simulated data with a controlled number of outliers.
Type-A datasets contain outliers that are clustered together, and generally distant from the rest of the data points.
In contrast, Type-B datasets contain outliers that are more evenly distributed.
Type-A and Type-B sets both include 12 datasets with different number of samples $n=\{60,100,200,500\}$ and features $m=\{2,5,10\}$.
Finally, for all of the considered benchmark datasets, \citet{Brooks2011} obtained five different instances with different random data points.
Overall, this gives us $(11+12+12) \times 5 = 175$ instances to evaluate SVM-HML solution methods. Additionally, for each of these instances, we will consider $C \in \{1,10,100,1000,10000\}$ for the penalty factor as done in~\cite{Brooks2011}, giving a total of 875 instances for each algorithm.

\begin{table}[htbp]
\centering
	\begin{tabular}{l@{\hspace*{1.5cm}}cc}
		\toprule
		Name &
		$n$ & 
		$m$ \\
		\midrule
		adult & 400 & 77\\
		australian & 366 & 45\\
		breast & 383 & 9\\
		bupa & 193 & 6\\
		german & 400 & 24\\
		heart & 152 & 20\\
		ionosphere & 196 & 33\\
		pima & 400 & 8\\
		sonar & 116 & 60\\
		wdbc & 319 & 30\\
		wpbc & 108 & 30\\
		\bottomrule
	\end{tabular}
	\caption{Characteristics of the UCI datasets}
    \label{tab:uci-instances}
\end{table}

All the algorithms considered in this study have been implemented in C++ and use the CPLEX 12.9 callable library. The experiments were run on an Intel Xeon E5-2620 2.1 GHz processor machine with 128 GB of RAM and CentOS Linux 7 (Core) operating system. All the source code and scripts needed to reproduce these experiments are provided at \url{https://github.com/vidalt/Hard-Margin-SVM}.

\subsection{Performance of CB-SVM-HML}
\label{subsec:first-experiments}

In the first set of experiments, we compare the results of the proposed CB-SVM-HML with those of the baseline algorithm of~\citet{Brooks2011}. We use the same experimental conditions, and therefore run each algorithm until a time limit of $T_\textsc{Max} = 600$\,seconds for each instance and value of $C$.
The time dedicated to the separation of \gls{CB} cuts in CB-SVM-HML is set to $T_\textsc{B} = T_\textsc{S} = 30$\,seconds. In other words, $5\%$ of the total time is dedicated to the separation of \gls{CB} on subsets of samples, and 5\% of the time on the complete problem. As shown in our sensitivity analyses in Section~\ref{subsec:second-experiments}, this amount of time already permits to separate a large number of cuts. Allocating more computational time for cut separation did not further improve the overall search process.

Tables~\ref{tab:results-uci-set}~to~\ref{tab:results-set-b-dim} report, for each algorithm over all $C$ values, the number of instances solved to optimality ({\#\,Opts}), the average optimality gap ({Gap\,(\%)}), and the average computational time in seconds ({Avg~Time}). In the last line, {Overall} provides the sum of all values for columns {\# Opts}, and the average of all values for columns {Gap (\%)} and {Avg Time}. In Tables~\ref{tab:results-set-a} and \ref{tab:results-set-b}, the results of the instance of Type A and Type B are aggregated per number of samples $n \in \{60,100,200,500\}$, whereas in Tables~\ref{tab:results-set-a-dim} and \ref{tab:results-set-b-dim}, the results are aggregated according to the dimension of the feature space $m \in \{2,5,10\}$. The detailed results for each instance are additionally provided in the same repository as the source code.\\

\begin{table}[htbp]
	\centering
	\begin{tabular}{ccccccccc}
		\toprule&&
		\multicolumn{3}{c}{\citet{Brooks2011}}&&\multicolumn{3}{c}{CB-SVM-HML}\\
		 &&\# Opts & Gap (\%) & Avg Time && \# Opts & Gap (\%)&Avg Time \\ \midrule
		 Overall&&149/275&33.32&295.31&&152/275&20.70&280.17\\  \bottomrule
	\end{tabular}
	\caption{Performance comparison on the UCI instances}
	\label{tab:results-uci-set}
\end{table}

\begin{table}[htbp]
	\centering
\begin{tabular}{ccccccccc}
		\toprule&&
		\multicolumn{3}{c}{\citet{Brooks2011}}&&\multicolumn{3}{c}{CB-SVM-HML}\\
		 $n$&&\# Opts & Gap (\%) & Avg Time && \# Opts & Gap (\%)&Avg Time \\ \midrule
		60&&75/75&0.00&2.87&&75/75&0.00&32.83 \\
		100&&42/75&11.66&325.71&&61/75&4.26&170.76\\
		200&&4/75&56.83&582.13&&32/75&18.21&395.30\\
		500&&0/75&88.17&600.00&&11/75&36.05&542.28\\ 
		\midrule
        Overall&&121/300&39.16&377.33&&179/300&14.63&285.29\\
		\bottomrule
	\end{tabular}
	\caption{Performance comparison on the Type-A instances -- grouped by number of samples $n$}
	\label{tab:results-set-a}
\end{table}

\begin{table}[htbp]
	\centering
	\begin{tabular}{ccccccccc}
		\toprule&&
		\multicolumn{3}{c}{\citet{Brooks2011}}&&\multicolumn{3}{c}{CB-SVM-HML}\\
		 $n$&&\# Opts & Gap (\%) & Avg Time && \# Opts & Gap (\%)&Avg Time \\ \midrule
		60&&74/75&0.28&72.01&&73/75&0.45&63.87 \\
		100&&26/75&26.63&404.67&&49/75&8.55&266.52 \\
		200&&1/75&65.59&595.16&&24/75&22.51&437.02 \\
		500&&0/75&90.76&600.00&&11/75&40.01&529.61
		\\ \midrule
        Overall&&101/300&45.82&417.61&&157/300&17.88&324.26\\
		\bottomrule
	\end{tabular}
	\caption{Performance comparison on the Type-B instances -- grouped by number of samples $n$}
	\label{tab:results-set-b}
\end{table}

\begin{table}[htbp]
	\centering
	\begin{tabular}{ccccccccc}
		\toprule&&
		\multicolumn{3}{c}{\citet{Brooks2011}}&&\multicolumn{3}{c}{CB-SVM-HML}\\
		 $m$&&\# Opts & Gap (\%) & Avg Time && \# Opts & Gap (\%)&Avg Time \\ \midrule
		2&&54/100&28.15&297.02&&86/100&1.84&145.27\\
		5&&40/100&40.37&388.24&&52/100&13.87&322.75\\
		10&&27/100&48.98&446.72&&41/100&28.18&387.85\\
		\midrule
        Overall&&121/300&39.16&377.33&&179/300&14.63&285.29\\
		\bottomrule
	\end{tabular}
	\caption{Performance comparison on the Type-A instances -- grouped by number of features $m$}
	\label{tab:results-set-a-dim}
\end{table}

\begin{table}[htbp]
	\centering
	\begin{tabular}{ccccccccc}
		\toprule&&
		\multicolumn{3}{c}{\citet{Brooks2011}}&&\multicolumn{3}{c}{CB-SVM-HML}\\
		 $m$&&\# Opts & Gap (\%) & Avg Time && \# Opts & Gap (\%)&Avg Time \\ \midrule
		2&&51/100&32.14&305.96&&83/100&1.60&148.83\\
		5&&26/100&48.06&455.49&&47/100&16.13&358.51\\
		10&&24/100&57.26&491.38&&27/100&35.90&465.42\\
		\midrule
        Overall&&101/300&45.82&417.61&&157/300&17.88&324.26\\
		\bottomrule
	\end{tabular}
	\caption{Performance comparison on the Type-B instances -- grouped by number of features $m$}
	\label{tab:results-set-b-dim}
\end{table}

As seen in these experiments, CB-SVM-HML generally achieves better performance than the baseline algorithm of \citet{Brooks2011}.
In general, it solves more instances to optimality with the same time limit (488/875 compared to 371/875) and achieves smaller optimality gaps (17.65\% on average compared to 39.61\%). CB-SVM-HML also found optimal solutions for some instances with 500 samples. Instances of this size could not be solved to proven optimality by previous approaches. Observing the results for instances with a different number of features~$m$, we see that CB-SVM-HML performs especially well on low-dimensional problems (i.e. when $m \in \{2,5\}$) since MIS are generally smaller in this regime, leading to stronger \gls{CB} cuts involving fewer variables. Generally, our method improved upon the baseline for all values of~$m$.

In terms of computational time, CB-SVM-HML achieves optimality or attains smaller gaps faster than the baseline algorithm on all instances except those with $n=60$ samples. For those small instances, the difference of performance comes from our parametrization choices: the current cut-separation algorithm uses at least 5\% of the time (30 seconds) separating cuts on randomly-generated subproblems including different subsets of the samples. As a consequence, the method's computational time is bounded below by 30 seconds, whereas the baseline algorithm sometimes solves the complete problem in shorter time for very small instances. One way to reduce computational effort in those cases could involve using a variable separation time that depends on the size of the instance, or setting a limit on the number of subproblems considered in the sampling phase.

\begin{figure}[htbp]
    \centering
    \includegraphics{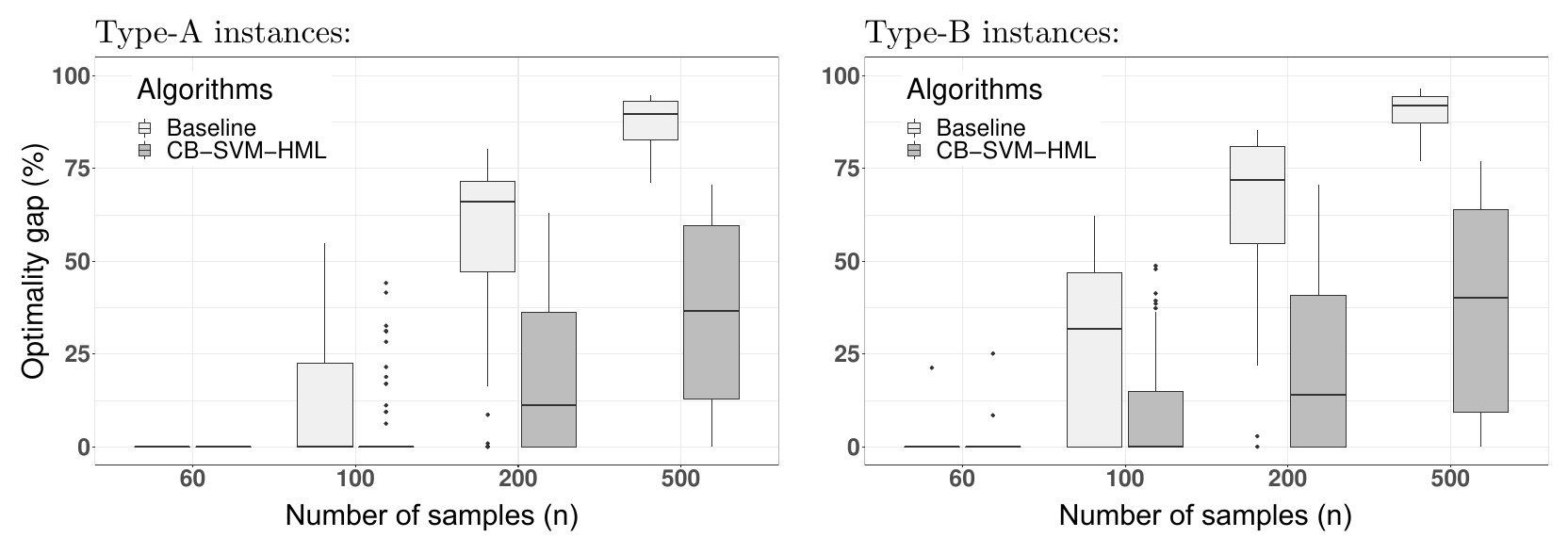} 
    \caption{Optimality gaps on Type-A and Type-B instances as a function of the number of samples}
    \label{fig:boxplot-ab-size-grouped-optimality-gap}
\end{figure}

We complete this analysis in Figure~\ref{fig:boxplot-ab-size-grouped-optimality-gap} with a fine-grained study of the optimality gaps of the methods as a function of the number of samples $n \in \{60,100,200,500\}$ in Type-A and Type-B instances. For each value of $n$ and for each method, we represent the optimality gaps (over the $75$ instances) as a boxplot. The boxes indicate the first and third quartile, and the whiskers extend to 1.5 times the interquartile range. Outliers that extend beyond this range are depicted as small dots.

As can be seen, CB-SVM-HML achieves better optimality gaps than the baseline algorithm for all values of $n$. Especially when $n=500$, the baseline method terminates with optimality gaps as high as $85\%$ in most cases, whereas CB-SVM-HML achieves much smaller gaps. This confirms the positive impact of the \gls{CB} cuts, which significantly tighten the formulation and decrease the optimality gap.

\subsection{Impact of the time dedicated to cut separation}
\label{subsec:second-experiments}

\begin{figure}[!htbp]
    \centering
    \includegraphics[width=0.95\textwidth]{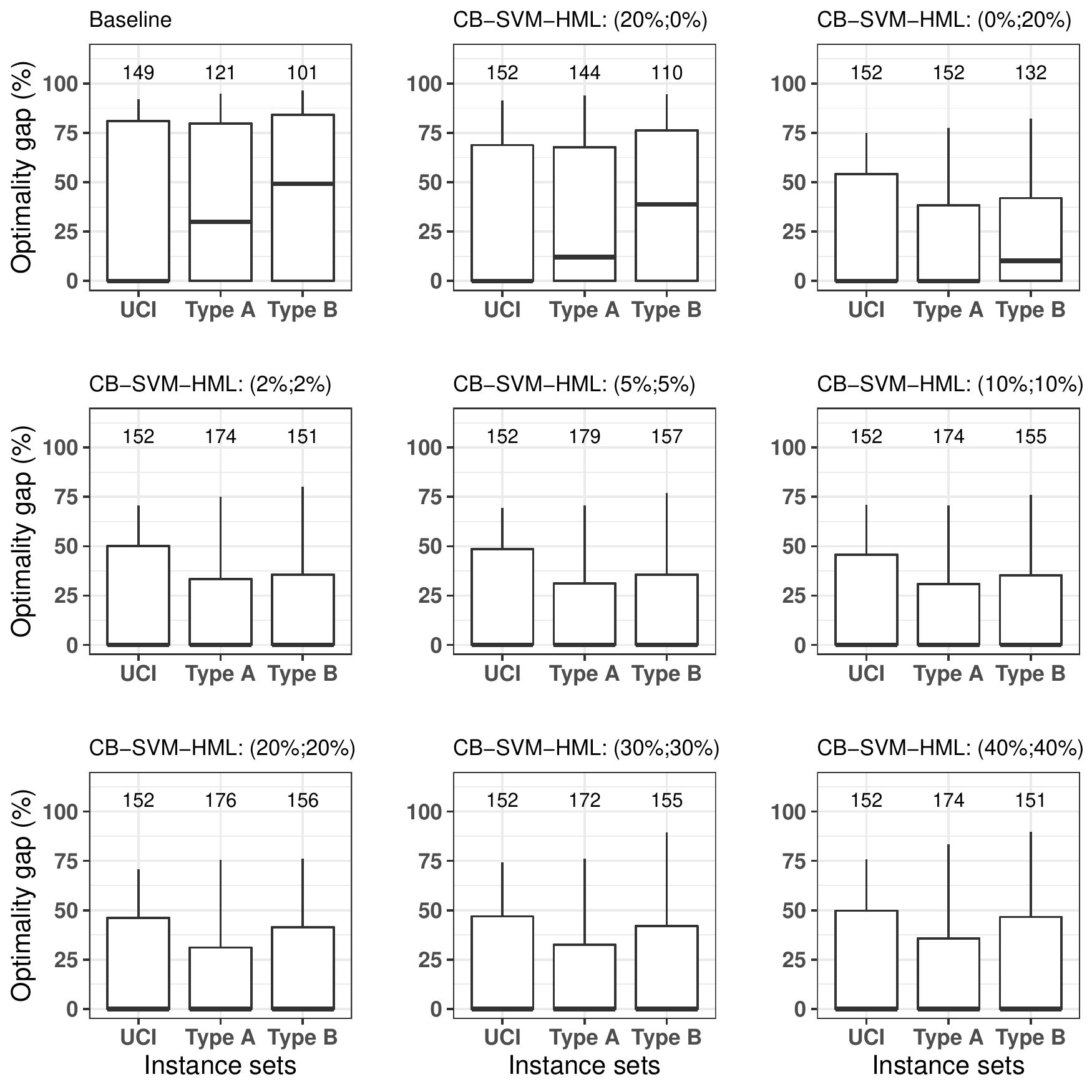}
    \caption{Optimality gaps achieved by CB-SVM-HML with different values of $T_{B}$ and $T_{S}$}
  \label{fig:boxplot-impact-of-cb-cuts}
\end{figure}

Finally, we measure the impact of the amount of computational effort dedicated to separating the \gls{CB} cuts on solution quality. We therefore compare eight different configurations of CB-SVM-HML with different time budgets for $T_\textsc{S}$ and $T_\textsc{B}$ to evaluate the impact of \gls{CB}-cuts separation on randomized sample subsets (during $T_\textsc{S}$) as well as on the complete problem (during $T_\textsc{B}$). For all configurations, we fix the total time limit to $T_\textsc{Max}$ to 600\,seconds.
Figure~\ref{fig:boxplot-impact-of-cb-cuts} shows the results of this experiment as boxplots, where each plot corresponds to the optimality gaps obtained by a method configuration on an instance set. Additionally, we indicate the number of optimal solutions found by each configuration on top of each boxplot.

As can be seen, the configuration $(5\%;5\%)$ (which is the \emph{reference configuration} used in Section~\ref{subsec:first-experiments}) achieved the best performance among all the considered configurations. Moreover, all the configurations with combinatorial Benders' cuts achieved better gaps and number of optimal solutions than the baseline algorithm of~\citet{Brooks2011}. Comparing configuration $(T_\textsc{S},T_\textsc{B}) = (10\%,10\%)$ with configurations $(20\%,0\%)$ and $(0\%,20\%)$, we notice that allocating the complete separation-time budget to the complete problem or to the randomized subproblems is not as effective as dividing the available time between these two approaches. Finally, allocating a larger amount of time to the separation process, as in configurations $(30\%,30\%)$ and $(40\%,40\%)$, also leads to a performance deterioration since there is a larger number of cuts, which leads to a heavier model, such that the remaining time to solve Step~3 is insufficient to obtain a good solution quality.

As seen in Table~\ref{tab:statistical-test}, the previous observations are confirmed at $0.05$ significance level by paired-samples Wilcoxon tests. All versions of CB-SVM-HML obtained optimality gaps which were significantly smaller than the baseline method of \citet{Brooks2011} without cuts. Moreover, the configuration of CB-SVM-HML with  $(T_\textsc{S},T_\textsc{B}) = (5\%,5\%)$ obtained better results than all configurations, except $(20\%,20\%)$ for which no statistical difference was observed.

\begin{table}[!htbp]
\centering
\caption{Significance results of paired-samples Wilcoxon tests}
\label{tab:statistical-test}
\renewcommand{\arraystretch}{1.25}
\setlength{\tabcolsep}{5pt} 
\scalebox{0.8}
{
 \begin{tabular}{lccccccccc} 
 \toprule
Configuration &&(20\%,0\%)& (0\%,20\%) & (2\%,2\%) & (5\%,5\%) & (10\%,10\%) & (20\%,20\%) & (30\%,30\%) & (40\%,40\%) \\
 \midrule
vs&p&$<$0.001&$<$0.001&$<$0.001&$<$0.001&$<$0.001&$<$0.001&$<$0.001&$<$0.001\\
Baseline~\citep{Brooks2011}&sign&\cmark&\cmark&\cmark&\cmark&\cmark&\cmark&\cmark&\cmark\\
 \midrule
vs&p&$<$0.001&$<$0.001&$<$0.001&-- &$<$0.001&0.796&$<$0.001&$<$0.001\\
(5\%,5\%)&sign&\cmark&\cmark&\cmark& -- &\cmark&&\cmark&\cmark\\
 \bottomrule
 \end{tabular}
}
\end{table}

\pagebreak
\section{Conclusion}
\label{section:conclusions}

In this study, we have introduced CB-SVM-HML, a mixed integer programming approach based on combinatorial Benders' cuts for optimally training the \gls{SVM-HML}. CB-SVM-HML operates by (i) generating an initial heuristic solution of the SVM-HML to obtain an upper bound on $||w||$, (ii) using this bound to define separation subproblems that permit to separate \gls{CB} cuts, and finally (iii) solving the SVM-HML with these additional cuts. Through extensive experiments, we observed that this methodology permits substantial advances in the solution of the SVM-HML, increasing our ability to achieve optimal solutions and small optimality gaps. Our sensitivity analyses show that additionally separating \gls{CB} cuts on small randomized subproblems with fewer samples also permits substantial performance improvements.

The findings of our study open many research avenues. First, we suggest pursuing methodological developments on mixed-integer programming strategies for the SVM-HML. Indeed, non-convex separation problems such as the SVM-HML with natural ``big-M'' MILP formulations are archetypal in many classification models (see, e.g., \citep{Carrizosa2020survey,Elizondo2006,Florio2022a,Freville2010,Hanafi2011,Murthy1994}), such that new developments realized on one problem can trigger significant advances for the others.
Next, while this study focuses on algorithms capable of proving optimality, research is still needed on efficient heuristics that can scale up to much larger data sets. Arguably, there is a need for both optimal algorithms and heuristics, since known optimal solutions give critical information regarding our ability to solve training problems reliably. Moreover, optimal or near-optimal solutions permit us to properly assess learning models without confounding factors due to the possible errors and instabilities of the training algorithms \citep{Gribel2019}. Finally, from a more general viewpoint, combinatorial optimization techniques can play an essential role in many other learning tasks and models. Especially given the rising concerns over equity, privacy, and transparency, sophisticated optimization strategies become necessary for challenging tasks related to model compression, training, and explanations, among others \citep{Forel2022,Parmentier2021c,Serra2020,Vidal2020a}.

\section*{Acknowledgments}

This research has been supported by CAPES-PROEX [grant number 88887.214468/2018-00], CNPq [grant number 308528/2018-2], and FAPERJ [grant number E-26/202.790/2019] in Brazil. This support is gratefully acknowledged.


\end{document}